\documentclass[…]{article}

\usepackage[preprint]{neurips_2025}
\usepackage[utf8]{inputenc} 
\usepackage[T1]{fontenc}    
\usepackage{hyperref}       
\usepackage{url}            
\usepackage{booktabs}       
\usepackage{amsfonts}       
\usepackage{nicefrac}       
\usepackage{microtype}      
\usepackage{xcolor}         
\usepackage{tcolorbox}
\usepackage{booktabs}
\usepackage{multirow}
\usepackage{tabularx}
\usepackage{caption}
\usepackage{amsmath}
\usepackage{amssymb}
\usepackage{wrapfig}
\usepackage{xspace}
\usepackage[utf8]{inputenc} 
\usepackage[T1]{fontenc}    
\usepackage{hyperref}       
\usepackage{url}            
\usepackage{booktabs}       
\usepackage{amsfonts}       
\usepackage{nicefrac}       
\usepackage{microtype}      
\usepackage{xcolor}         
\usepackage{graphicx}
\usepackage{multirow} 
\usepackage{subcaption}
\usepackage{cleveref}
\usepackage{enumitem}
\usepackage[numbers,sort&compress]{natbib}

\newcommand{\met}{\textit{SEM}\xspace}
\hypersetup{
  colorlinks,
  linkcolor={blue!70!green},
  citecolor={green!70!blue},
  urlcolor={orange!70!red}
}
\usepackage{graphicx}
\usepackage{subcaption}

\newcommand{\mypara}[1]{\smallskip\noindent{\bf {#1}.}\xspace}

\newtcolorbox{cotbox}[1][]{
    colback=maincolor!10,
    colframe=maincolor,
    width=\columnwidth,
    fonttitle=\bfseries,
    coltitle=white,
    arc=1mm,
    auto outer arc,
    left=4pt,
    right=4pt,
    breakable,
    title=#1,
}

\title{\textbf{SEM}: \textsc{Reinforcement Learning for \underline{S}earch-\underline{E}fficient Large Language \underline{M}odels}}

\author{%
  Zeyang Sha \\
  Ant Group \\
    \texttt{shazeyang.szy@antgroup.com} \\
  \And
  Shiwen Cui\\
  Ant Group \\
    \texttt{donn.csw@antgroup.com} \\
  \AND
  Weiqiang Wang \\
  Ant Group \\
    \texttt{weiqiang.wwq@antgroup.com} \\
}

\begin{document}

\maketitle

\begin{abstract}
Recent advancements in Large Language Models (LLMs) have demonstrated their capabilities not only in reasoning but also in invoking external tools, particularly search engines.
However, teaching models to discern when to invoke search and when to rely on their internal knowledge remains a significant challenge. 
Existing reinforcement learning approaches often lead to redundant search behaviors, resulting in inefficiencies and over-cost. 
In this paper, we propose \met, a novel post-training reinforcement learning framework that explicitly trains LLMs to optimize search usage.
By constructing a balanced dataset combining Musique and MMLU, we create scenarios where the model must learn to distinguish between questions it can answer directly and those requiring external retrieval.
We design a structured reasoning template and employ Group Relative Policy Optimization (GRPO) to post-train the model’s search behaviors. 
Our reward function encourages accurate answering without unnecessary search while promoting effective retrieval when needed. 
Experimental results demonstrate that our method significantly reduces redundant search operations while maintaining or improving answer accuracy across multiple challenging benchmarks.
This framework advances the model's reasoning efficiency and extends its capability to judiciously leverage external knowledge. 

\end{abstract}

\section{Introduction}

Recent advancements in Large Language Models (LLMs) have underscored the considerable benefits of incorporating extended reasoning processes to enhance their performance on complex tasks~\cite{plaat2024reasoning,xu2025towards}.
Beyond their reasoning capabilities, LLMs have also demonstrated a surprising aptitude for tool invocation~\cite{DBLP:journals/corr/abs-2409-18807,DBLP:journals/csur/QinHLCDCZZHXHFSWQTZLSXZ25,DBLP:conf/nips/YangSLZGLS23,DBLP:conf/naacl/QiaoGLJC024}.
By explicitly instructing the model through prompts on when and how to invoke external tools, it becomes capable of performing tasks beyond the limits of pure linguistic reasoning.

Among various tools, the search functionality stands out as particularly essential~\cite{deepresearch}.
When confronted with uncertain or unfamiliar questions, models can leverage search interfaces to retrieve relevant information, subsequently using the acquired data to generate more accurate and contextually precise responses.

Teaching models to effectively utilize search functions has presented significant challenges. 
The most straightforward method involves embedding explicit instructions within the context prompts~\cite{DBLP:conf/acl/TrivediBKS23,DBLP:conf/emnlp/ShaoGSHDC23, DBLP:journals/corr/abs-2501-05366}. 
If a model has robust contextual understanding, it can efficiently learn and apply these instructions, invoking appropriate tools when necessary. 
However, models frequently encounter difficulties in mastering sophisticated search behaviors, particularly in recognizing errors from initial searches and initiating subsequent searches—an issue commonly observed during iterative search interactions.

Previous research has illustrated the potential of reinforcement learning in training models to optimize their search behaviors~\cite{DBLP:journals/corr/abs-2503-19470, feng2025retoolreinforcementlearningstrategic, zheng2025deepresearcherscalingdeepresearch}.
By incorporating reward mechanisms tied to the efficacy of searches, models progressively enhance their understanding and utilization of search tools. 
Nonetheless, this approach has noticeable limitations, notably that models often execute searches unnecessarily, irrespective of the actual need.

As demonstrated in \autoref{tab:easy-case-study}, even for a trivial question like ``1+1=?'', the model redundantly performs multiple unnecessary searches, such as queries on ``the basic principle of addition.''
Clearly, there is a pressing need to optimize how models discern when a search is truly necessary, preventing wasteful use of resources and ensuring efficiency in their reasoning processes.

\subsection{Our Contribution}
Addressing these challenges, we introduce a novel post-training reinforcement learning framework, \met, designed specifically to teach models to distinguish when to invoke search and when it is unnecessary.
Specifically, if the model is confident in its understanding of a question, it directly outputs the answer without invoking search tools. 
Conversely, if the model is uncertain, it will initiate a search to acquire relevant context, thereby enhancing its comprehension and response accuracy.

To equip the model with effective search‐awareness, we construct a balanced dataset comprising two equal portions: one in which the model already knows the correct answers and one in which it does not. During training, we require the model to generate its responses using explicit <think> and <answer> annotations.
Whenever the model’s initial <answer> is correct, we impose a penalty on any subsequent attempt to invoke the search tool. In contrast, if the initial <answer> is incorrect, we provide a positive reward for issuing a <search> query. In this latter scenario, the model is expected first to emit a <search> request, retrieve relevant information, and then produce a refined <answer> based on the newly acquired knowledge.

This carefully structured reinforcement pipeline enables the model to progressively sharpen its judgment on the necessity of searches, significantly reducing redundant actions and enhancing response precision.
Extensive evaluations affirm that our proposed framework substantially improves the model's efficiency and effectiveness, empowering it to leverage search tools optimally, especially in complex and uncertain scenarios.

\mypara{Implication}
The proposed \met significantly enhances the efficiency and accuracy of LLMs in utilizing external search tools. 
By explicitly training models to discern when external retrieval is necessary, this approach substantially reduces redundant search behaviors, thereby optimizing resource usage. 
Moreover, equipping models with robust search-awareness extends their reasoning capabilities, enabling them to handle a broader range of complex and uncertain questions effectively. 
This advancement not only improves the practical applicability of LLMs in real-world scenarios but also sets a foundation for more sophisticated integrations of external tools, paving the way for future developments in interactive and context-aware AI systems.

\begin{figure}[htbp]
  \centering
\includegraphics[width=\textwidth]{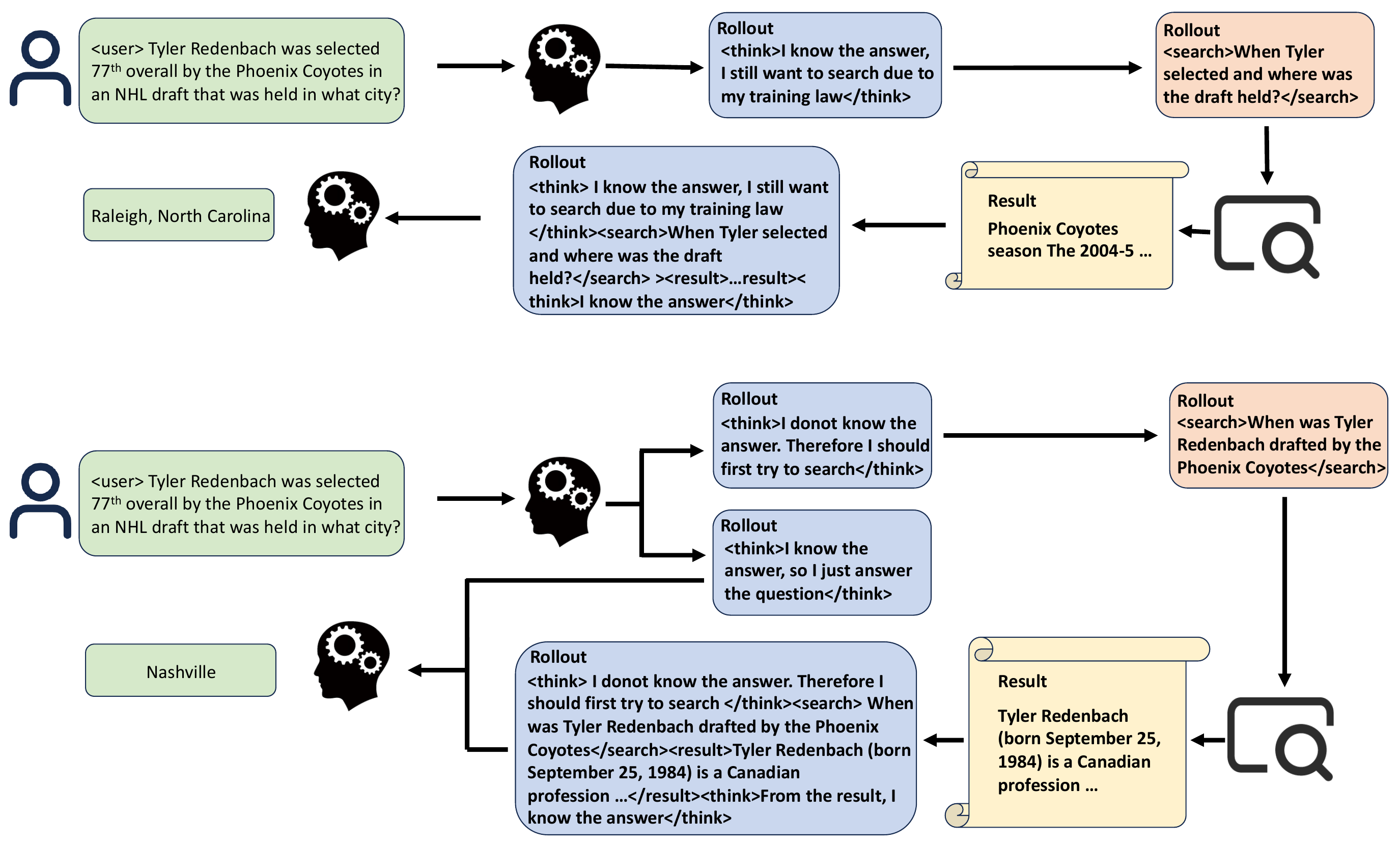}
  \caption{Comparison of the previous method and \met.}
  \label{fig:method}
\end{figure}

\section{Method}

To effectively integrate external search capabilities into the model’s intrinsic reasoning mechanisms, we employ reinforcement learning to post-train the base model. 
The comprehensive methodology is illustrated in \autoref{fig:method}.

\subsection{Dataset Preparation}
\label{sec:dataset_pre}

The primary goal of \met is to equip the model with the ability to make intelligent use of external search tools in order to improve the accuracy and relevance of its responses to user queries. 
Rather than relying solely on its internal knowledge, the model should learn when to retrieve additional information through search and how to incorporate the retrieved content into its final answer effectively.

To develop this capability, the first critical step is to construct a training dataset that explicitly reflects the need for such behavior. 
Specifically, we aim to provide a clear distinction between questions that the model is likely to answer correctly using its existing knowledge and those for which it lacks sufficient information and would benefit from a search.

To this end, we combine two complementary datasets—Musique~\cite{trivedi2021musique} and MMLU~\cite{DBLP:conf/iclr/HendrycksBBZMSS21}—to form the training corpus. 
Musique primarily consists of multi-hop, fact-based questions that often go beyond the model's pretraining knowledge, making them ideal candidates for demonstrating the value of search. 
In contrast, MMLU includes a broad range of academic and professional exam questions that are generally well-covered in existing training corpora, and thus, are typically answerable without search.

By integrating these two datasets, we establish a balanced training distribution that includes both ``known'' and ``unknown'' questions.
This balanced composition enables us to design a reinforcement learning framework where the model is rewarded differently based on context: it receives direct positive feedback for answering known questions correctly without invoking search, while it is incentivized to use search strategically in cases where its initial response is insufficient or incorrect.

This approach not only encourages the model to recognize its knowledge boundaries but also helps it develop a decision-making process around when and how to invoke search.
Over time, the model learns to optimize for both answer quality and computational efficiency by selectively engaging the search module only when necessary.

\subsection{Reward Policy}

We implement a carefully structured reward policy to teach the model effective search reasoning, utilizing the Group Relative Policy Optimization (GRPO) framework for optimization.

\mypara{Group Relative Policy Optimization~\cite{DBLP:journals/corr/abs-2402-03300}}  
To effectively teach the model when and how to utilize external tools such as search engines, we adopt a Group Relative Policy Optimization (GRPO) framework. 
Instead of applying a uniform reward across all trajectories, GRPO considers the relative quality of model outputs within the same query group. 
This encourages the model to produce the best possible reasoning chain for a given input, even if the final answer is correct in multiple cases. 

\mypara{Reward Modeling}  
Our reward function is designed to simultaneously encourage correct reasoning without unnecessary tool invocation and incentivize effective search usage when required. Concretely, the model is rewarded for:  
(1) correctly predicting the answer without relying on search when its internal knowledge suffices,  
(2) using search judiciously when the question is beyond its knowledge scope, and  
(3) adhering to a strict response format that includes <think>, <answer>, <search>, <result> tags in proper order.
We show the reward formula in \autoref{equ:reward}.

Specifically, we first extract all answers enclosed in the <answer> tag and evaluate their correctness using an F1 score based on token-level overlap with the ground truth. 
If the first answer achieves a high F1 score (above a predefined threshold), but the model still invokes search or produces redundant reasoning steps, it is penalized for unnecessary exploration. 
Otherwise, when the first answer is incorrect, the model must engage in search and generate a second answer, which is then evaluated for correctness. 
Invalid formatting or improper tag ordering results in zero reward.

\begin{equation}
\mathcal{R}
= f \Bigl[
\mathbf{1}\{F_{1}(a_{1})\ge\tau \wedge s=0 \wedge t=1\}\,F_{1}(a_{1})
\;+\;
\mathbf{1}\{F_{1}(a_{1})<\tau \wedge u=1\}\,F_{1}(a_{2})
\Bigr].
\label{equ:reward}
\end{equation}

\noindent where
\begin{align*}
f &\in \{0,1\}, &&\text{valid structure indicator},\\
s &\in \{0,1\}, &&\text{search-invoked indicator},\\
t &\in \{0,1\}, &&\text{single think/answer indicator},\\
u &\in \{0,1\}, &&\text{valid search–result format indicator},\\
\tau &\in \mathbb{R}, &&\text{confidence threshold},\\
& \mathbf{1}\{\cdot\} && \text{indicator function}.
\end{align*}

\mypara{Rollout with Search}  
During rollout, the model generates its complete reasoning trajectory in a structured template that includes optional search. The search invocation is treated as an intermediate sub-action between two phases of reasoning. If a search is triggered, the model must produce a <search> query, followed by the <result> retrieved from the external source, and finally update its belief state before issuing the final <answer>. This design allows us to explicitly assess the impact of search and isolate its contribution to the accuracy of the final output.

\subsection{Training Template}

To standardize the reasoning and search process, we define a consistent response format used throughout training. Each model output follows the template:

\begin{verbatim}
<think> initial reasoning </think>
<answer> preliminary answer </answer>
<search> search query (if any) </search>
<result> retrieved result </result>
<think> updated reasoning based on retrieved info </think>
<answer> final answer </answer>
\end{verbatim}

This structured format allows robust parsing and evaluation during reward computation. 
It also supports modular supervision, enabling us to provide targeted feedback on both the reasoning quality and the utility of search. 
Models are trained to optimize for both accuracy and minimal, justified usage of external tools, promoting a balance between confidence and curiosity.

\section{Experiments}

\subsection{Experimental Setup}

\mypara{Datasets}
As we have stated in \autoref{sec:dataset_pre}, to enable the model to learn when it knows the answer and when it does not, we build the training dataset combining Musique and MMLU.
Specifically, most of the questions in Musique are unfamiliar to the model, and the use of retrieval-augmented generation (RAG)~\cite{DBLP:conf/nips/LewisPPPKGKLYR020,DBLP:conf/iclr/AsaiWWSH24,DBLP:conf/iclr/YoranWRB24} significantly improves its ability to answer these questions.
In contrast, the questions in MMLU are generally within the model’s existing knowledge, making an external search unnecessary.

After training, we evaluate the model on Musique~\cite{trivedi2021musique} and HotpotQA~\cite{DBLP:conf/emnlp/Yang0ZBCSM18}, where the questions are challenging for LLMs, as well as on MMLU~\cite{DBLP:conf/iclr/HendrycksBBZMSS21} and GSM8k~\cite{DBLP:journals/corr/abs-2110-14168}, which consists of logic math problems that typically do not require search.

\mypara{Metrics}
We consider three metrics: Exact Match(EM), LLM as a Judger(LJ), and Search Ratio(SR) to measure the results of the trained model
We compute the EM by measuring the percentage of examples for which the model's final answer exactly matches one of the ground-truth answers.
However, EM metric is too hard to measure the accuracy of the model answer due to the fact that sometimes, the model's answer is right but only a few words are different from the ground truth.
In this case, we also use LLMs to determine whether the answer is correct or not.
We take advantage of deepseek-671B AWQ~\cite{deepseekai2025deepseekr1incentivizingreasoningcapability} as a judge.
Note that for datasets like MMLU or GSM8k, there is no need to use LJ as the model can always answer the exact right number or choices from the given options.
Moreover, we also consider the SR as one of the metrics.
We emphasize that in different cases, the SR should be different.
For datasets like Musique and HotpotQA, a higher SR is better as the questions are unknown for the models.
For other datasets like MMLU and GSM8k, the lower SR is better due to the fact that these questions are all logical reasoning questions that do not rely on external knowledge but the internal ability of the model.

\mypara{Implementation}
We implement our training framework based on ReSearch~\cite{DBLP:journals/corr/abs-2503-19470}, Verl~\cite{DBLP:conf/eurosys/ShengZYWZZPL025}, and FlashRAG~\cite{DBLP:journals/corr/abs-2405-13576}
Note that we train the model for only 200 steps because, in the reinforcement learning setup, this number of updates is already sufficient to observe significant gains in performance.
We retrieve the information from the wiki18-100w.
We take advantage of Qwen models~\cite{yang2024qwen2} as our base models.
We use 8 A100 and set the batch size as 8.

\subsection{Main Results}

\begin{table}[ht]
  \centering
  \caption{Performance on HotpotQA and MuSiQue.}
  \label{tab:7b-14b-comparison-hotpot-musique}
  \begin{tabularx}{\textwidth}{l l *{3}{>{\centering\arraybackslash}X}}
    \toprule
    \textbf{Dataset}   & \textbf{Model}      & \textbf{EM}    & \textbf{LJ}    & \textbf{SR} \\
    \midrule
    \multirow{7}{*}{HotpotQA}
      & \textit{7B-Instruct} &                &                &                      \\
      & \quad Naive RAG      & 18.01          & 47.51          & 88.52\%              \\
      & \quad ReSearch       & 21.75          & 32.06          & 0.08\%               \\
      & \quad \met           & 35.84          & 61.67          & 97.54\%              \\
      \cmidrule(lr){2-5}
      & \textit{14B-Instruct}&                &                &                      \\
      & \quad Naive RAG      & 35.11          & 59.55          & 87.66\%              \\
      & \quad ReSearch       & 33.29          & 52.01          & 100.00\%             \\
      & \quad \met           & 40.42          & 58.43          & 98.77\%                    \\
    \midrule
    \multirow{7}{*}{MuSiQue}
      & \textit{7B-Instruct} &                &                &                      \\
      & \quad Naive RAG      & 7.19           & 26.69          & 90.57\%              \\
      & \quad ReSearch       & 6.08           & 11.67          & 0.12\%               \\
      & \quad \met           & 15.59          & 36.41          & 97.35\%              \\
      \cmidrule(lr){2-5}
      & \textit{14B-Instruct}&                &                &                      \\
      & \quad Naive RAG      & 13.52          & 30.74          & 82.33\%              \\
      & \quad ReSearch      & 14.43          & 29.42          & 100.00\%             \\
      & \quad \met           & 20.56              & 32.28              & 97.10\%                    \\
    \bottomrule
  \end{tabularx}
\end{table}

\begin{table}[ht]
  \centering
  \caption{Performance on MMLU and GSM8K.}
  \label{tab:7b-14b-comparison-mmlu-gsm8k}
  \begin{tabularx}{\textwidth}{l l *{2}{>{\centering\arraybackslash}X}}
    \toprule
    \textbf{Dataset}   & \textbf{Model}      & \textbf{EM}    & \textbf{SR} \\
    \midrule
    \multirow{7}{*}{MMLU}
      & \textit{7B-Instruct} &                &                      \\
      & \quad Naive RAG      & 12.48          & 47.98\%              \\
      & \quad ReSearch       & 69.84          & 0.00\%               \\
      & \quad \met           & 70.88          & 1.77\%               \\
      \cmidrule(lr){2-4}
      & \textit{14B-Instruct}&                &                      \\
      & \quad Naive RAG      & 70.49              & 11.74\%                    \\
      & \quad ReSearch       & 75.16          & 31.43\%              \\
      & \quad \met           & 75.62          & 0.11\%                    \\
    \midrule
    \multirow{7}{*}{GSM8K}
      & \textit{7B-Instruct} &                &                      \\
      & \quad Naive RAG      & 12.48          & 61.56\%              \\
      & \quad ReSearch       & 82.63          & 0.00\%               \\
      & \quad \met           & 71.79          & 14.63\%              \\
      \cmidrule(lr){2-4}
      & \textit{14B-Instruct}&                &                      \\
      & \quad Naive RAG      & 83.93              & 14.71\%                    \\
      & \quad ReSearch       & 50.41          & 55.19\%                    \\
      & \quad \met           & 79.37          & 0.76\%                    \\
    \bottomrule
  \end{tabularx}
\end{table}

We present the performance results of our experiments in \autoref{tab:7b-14b-comparison-hotpot-musique} and \autoref{tab:7b-14b-comparison-mmlu-gsm8k}. 
It is important to note that our ReSearch results differ from the original paper due to discrepancies in training datasets.

As demonstrated in the table, our proposed \met consistently demonstrates superior performance across all evaluated benchmarks.
On the HotpotQA dataset, our Qwen2.5-7B-Instruct model trained under \met achieves an Exact Match (EM) score of 35.84, significantly outperforming the Naive RAG approach, which attains an EM of only 18.01. 
This improvement indicates that our reinforcement learning (RL) framework effectively teaches the model when and how to perform external searches. 
Similar trends are observed on MuSiQue, where the EM score for our 7B-Instruct model (15.59) markedly surpasses the Naive RAG's 7.19, reinforcing the effectiveness of our search-optimized training.

For logic-based datasets such as MMLU, our method also excels at guiding the model to recognize when internal knowledge suffices, thereby avoiding unnecessary searches.
Specifically, the search ratio for our Qwen2.5-7B-Instruct model on MMLU is impressively low at 1.77\%, substantially less than that of both Naive RAG (47.98\%).
Note that the Qwen2.5-7B-Instruct model trained with ReSearch lacks the ability to perform search, resulting in a 0\% SR.
Remarkably, even without frequent search invocations, our model achieves a robust EM score of 70.88, whereas the Naive RAG model manages only 12.48. 
These results highlight the dual benefits of our proposed approach: enhancing search decision-making capabilities and significantly awakening the model's intrinsic reasoning and teaching-following abilities.

Conversely, ReSearch exhibits relatively weaker performance under our experimental setup.
This degradation is primarily due to the composition of our training dataset, which makes models trained with the ReSearch framework prone to gradient explosion.
As a result, these models either excessively rely on search or fail to utilize it effectively, ultimately leading to lower accuracy compared to our method.

The GSM8K results further illustrate these dynamics.
Note that the model is only trained on MMLU, which is totally different with the GSM8k.
However, the model trained under \met can still achieve great results without redundant search.
For instance, Qwen2.5-7B-Instruct trained under \met only invoke 0.76\% search during the all queries.
Moreover, the thinking process can still make the model maintain high accuracy on the math problems as Qwen-2.5-14B-Instruct can achieve 79.37\% EM, which is much higher than same model trained under ReSearch(50.41).

Overall, our results clearly indicate that the proposed \met significantly enhances the model's ability to discern when external information retrieval is beneficial, substantially improves its reasoning capabilities, and promotes adherence to structured response protocols.

\begin{figure}[htbp]
  \centering
  \includegraphics[width=\textwidth]{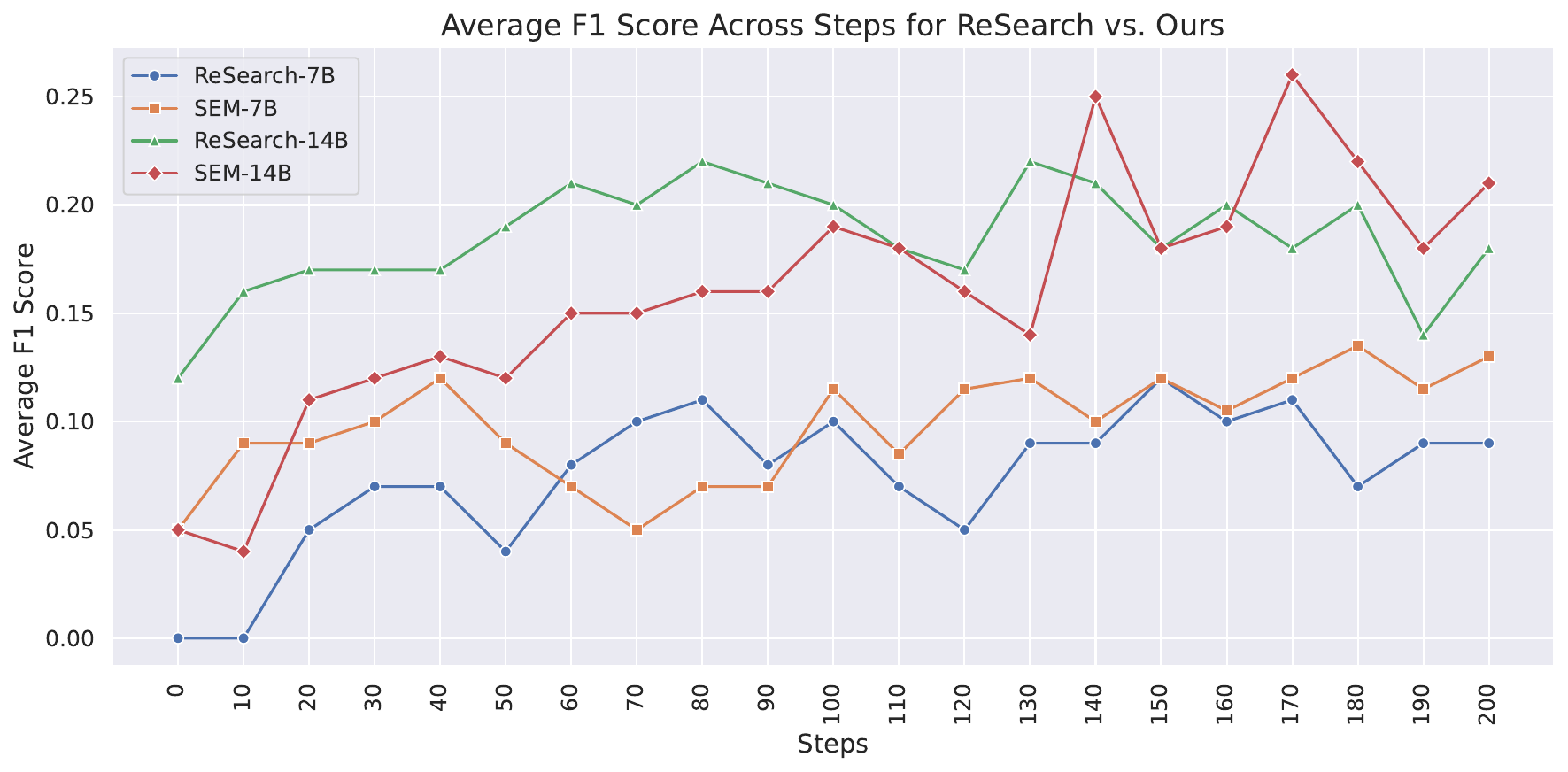}
  \caption{The performance of different models on different training steps.}
  \label{fig:accuracy_over_steps}
\end{figure}

\begin{table}[ht]
  \centering
  \caption{Case Studys for known Questions.}
  \label{tab:easy-case-study}
  \begin{tabularx}{\textwidth}{>{\bfseries}l X}
    \toprule
    \textbf{Questions} & 1+1=? \\
    \midrule
    ReSearch   &  <think> This is a straightforward mathematical question. I will first check if there is any need to search for information to solve this problem.</think><search> search query: What is the basic principle of addition in mathematics? </search><result>some result here</result><think> The information provided is about the rule of sum in combinatorics, which is not directly related to the question. However, the basic principle of addition is clearly stated in the arithmetic section. Since the question is about 1+1, I can directly solve this using the basic principle of addition.</think><search> search query: What is 1+1 in mathematics? </search><result>some result here</result><think> The information provided is about the concept of 1+1 in the context of infinite series and other mathematical contexts, but the question is asking for the simplest form of 1+1. I can directly use the basic principle of addition to solve this.</think><search> search query: What is the result of 1+1 in basic arithmetic? </search><result>some result here</result> <think> The information confirms that the simplest form of 1+1 in basic arithmetic is 2. I can now provide the answer.</think><answer> \verb|\boxed{2}| </answer> \\
    \addlinespace[1em]
    \met  & <think>The question is asking for the sum of 1+1. This is a basic arithmetic operation.</think> 
<answer>The answer is \verb|\boxed{2}.</answer>| \\
    \bottomrule
    \end{tabularx}
\end{table}

\begin{table}[ht]
  \centering
  \caption{Case Studys for Unknown Questions.}
  \label{tab:case-study}
  \begin{tabularx}{\textwidth}{>{\bfseries}l X}
    \toprule
    \textbf{Questions} & Tyler Redenbach was selected 77th overall by the Phoenix Coyotes in an NHL draft that was held in what city? \\
    \midrule
    ReSearch   & <think> To identify the NHL draft city in which Tyler Redenbach was selected, I will first need to ascertain the specific draft year he was chosen. I will then use that information to find the city where the draft was conducted.</think><search> search query: When was Tyler Redenbach drafted by the Phoenix Coyotes? </search><result>some result here</result><think> Now that I have the draft year, I can proceed to ascertain the location of the draft ceremony. I will now search for the city that hosted the 2003 NHL Entry Draft.</think><search> search query: Where was the 2003 NHL Entry Draft held? </search><result>some result here</result><think> Having found the location of the 2003 NHL Entry Draft, I can now ascertain the city in which Tyler Redenbach was drafted by the Phoenix Coyotes.</think><search> search query: Where was the 2003 NHL Entry Draft? </search><result>some result here</result><think> Now that I have confirmed the location of the 2003 NHL Entry Draft, I can conclude that Tyler Redenbach was drafted in Nashville, Tennessee.</think>
    <answer> \verb|\boxed{Nashville}| </answer> \\
    \addlinespace[1em]
    \met  & <think>To answer this, I need to find information about the NHL draft where Tyler Redenbach was selected 77th by the Phoenix Coyotes. I'll start by finding details about Tyler Redenbach and his draft information.</think> 
    <answer> The inital answer is \verb|\boxed{}|. </answer><search>Wikipedia search query: Tyler Redenbach NHL draft</search><result>some result here</result><think>From the information provided, we know Tyler Redenbach was drafted by the Phoenix Coyotes in 2003. Now, I need to find the location of the 2003 NHL Entry Draft.</think><search>Wikipedia search query: 2003 NHL Entry Draft location</search><result>some result here</result><think>The 2003 NHL Entry Draft, where Tyler Redenbach was selected, was held in Nashville, Tennessee. </think><answer>The Phoenix Coyotes selected Tyler Redenbach 77th overall in the 2003 NHL Entry Draft, which was held in Nashville, Tennessee. \verb|\boxed{Nashville, Tennessee}|</answer> \\
    \bottomrule
  \end{tabularx}
\end{table}

\subsection{Training Process}

We present the training dynamics of our models in \autoref{fig:accuracy_over_steps} to illustrate the effectiveness of the proposed framework.
The plotted curves represent the average F1 scores computed over 100 evaluation samples at every checkpoint, where an answer is considered correct if it achieves an F1 score of 1.0.
As shown in the figure, \met consistently outperforms the ReSearch baseline across both the 7B and 14B model sizes. 
Notably, the performance of our approach improves steadily over time, exhibiting a smoother and more stable learning trajectory.
In contrast, the ReSearch models suffer from larger fluctuations and slower gains in F1 score, particularly in the 7B setting. 
The advantage of our method becomes more pronounced in the 14B model, where it maintains a significantly higher F1 score throughout training. 
These results suggest that our framework not only accelerates convergence but also enhances the model’s ability to generalize more reliably as training progresses.

\subsection{Case Study}

To further highlight the effectiveness of the proposed framework, we present illustrative case studies in \autoref{tab:case-study}, showcasing the reasoning behaviors of models trained under ReSearch and \met.
All responses are generated under a unified prompting format that incorporates a system-defined reasoning template involving \texttt{<think>}, \texttt{<search>}, \texttt{<result>}, and \texttt{<answer>} tags.
This standardized structure ensures a fair comparison of model behaviors across different reasoning scenarios.

The first case in \autoref{tab:case-study} features a simple arithmetic question: \textit{``1+1=?''}. 
As this question lies well within the model’s internal knowledge, an ideal agent should answer it directly without invoking external search.
The ReSearch model, however, redundantly queries multiple times, demonstrating inefficient tool usage.
In contrast, our model recognizes the simplicity of the problem and immediately produces the correct answer without performing any unnecessary retrieval, showcasing a more efficient and targeted reasoning process.

The second example demonstrated in \autoref{tab:case-study} involves a fact-based open-domain question: \textit{``Tyler Redenbach was selected 77th overall by the Phoenix Coyotes in an NHL draft that was held in what city?''}. 
This query requires external factual knowledge beyond the model’s pretraining corpus.
The ReSearch model executes a multi-step search to first determine the draft year and then locate the corresponding host city, ultimately yielding the correct answer.
Our model exhibits similar multi-hop reasoning but accomplishes the task with fewer and more focused search operations, demonstrating improved retrieval efficiency and interpretability. 

These examples collectively underscore two key advantages of our approach: (1) the ability to avoid unnecessary retrieval for answerable questions, and (2) the capability to efficiently orchestrate multi-hop retrieval when external information is required. 
Such dynamic control over tool invocation is critical for enhancing both the interpretability and computational efficiency of tool-augmented language models.

\section{Related Work}

\subsection{Reinforcement Learning}

Reinforcement learning (RL)~\cite{DBLP:journals/nature/MnihKSRVBGRFOPB15,DBLP:conf/icml/WangSHHLF16,DBLP:journals/corr/ThomasB17,DBLP:conf/icml/MnihBMGLHSK16} has become a cornerstone in aligning large language models (LLMs) with human preferences and enhancing their reasoning capabilities. 
Proximal Policy Optimization(PPO)~\cite{DBLP:journals/corr/SchulmanWDRK17} is a widely adopted policy gradient method in RL that balances exploration and exploitation by limiting the deviation from the current policy during updates. 
Direct Preference Optimization (DPO)~\cite{DBLP:conf/nips/RafailovSMMEF23} offers a streamlined alternative to traditional RL approaches by directly optimizing the model's parameters based on human preference data. 
Unlike methods that require training a separate reward model, DPO simplifies the alignment process through a classification loss that encourages the model to prefer responses aligned with human preferences. 

Group Relative Policy Optimization (GRPO)~\cite{DBLP:journals/corr/abs-2402-03300} builds upon the foundations of PPO by introducing a group-based comparison mechanism. 
Instead of evaluating individual responses, GRPO assesses groups of outputs to derive a relative advantage, promoting more nuanced learning. 
This method has shown promise in enhancing the reasoning abilities of LLMs, particularly in complex tasks such as mathematical problem-solving~\cite{zhang2025grpoleaddifficultyawarereinforcementlearning}.

Collectively, these reinforcement learning methodologies contribute significantly to the post-training refinement of LLMs, ensuring that the models not only generate coherent text but also align closely with human expectations and demonstrate improved reasoning skills.

\subsection{Large Language Models as Agents}

Framing LLMs as autonomous agents capable of planning and executing multi-step reasoning has become an emerging paradigm~\cite{DBLP:journals/corr/abs-2503-21460,DBLP:journals/corr/abs-2309-14365,DBLP:journals/tmlr/SumersYN024,AutoAgent}. 
Agent frameworks such as AutoGPT~\cite{DBLP:journals/corr/abs-2306-02224} and LangChain~\cite{langchain} demonstrate how models can iteratively refine tasks, search information, and generate solutions. 
Recent work \cite{DBLP:journals/corr/abs-2503-19470, feng2025retoolreinforcementlearningstrategic, zheng2025deepresearcherscalingdeepresearch} emphasizes the importance of tool selection and usage timing. 
However, existing systems often rely on heuristics or fixed prompting strategies to manage tool invocation. 
In contrast, our method explicitly trains the model through RL to learn optimal tool usage patterns, enhancing both interpretability and performance.

\section{Conclusion}

In this work, we proposed a novel post-training reinforcement learning framework, \met, to optimize search behavior in large language models.
We first construct a balanced dataset that explicitly distinguishes between known and unknown questions, and designing a reward function that penalizes unnecessary search while encouraging effective retrieval.

Our experimental results demonstrate that we can significantly improve both the efficiency and performance of tool-augmented models.
Specifically, we train the model on the dataset combined by MuSiQue and MMLU and then evaluate the model on HotpotQA, MuSiQue, MMLU, and GSM8k.
Our results demonstrate that \met not only reduces redundant search operations but also enhances answer accuracy.
This work opens new directions for training intelligent and resource-efficient agents.

\newpage
\bibliographystyle{unsrt}
\bibliography{ref}

\begin{thebibliography}{10}

\bibitem{plaat2024reasoning}
Aske Plaat, Annie Wong, Suzan Verberne, Joost Broekens, Niki van Stein, and Thomas Back.
\newblock Reasoning with large language models, a survey.
\newblock {\em arXiv preprint arXiv:2407.11511}, 2024.

\bibitem{xu2025towards}
Fengli Xu, Qianyue Hao, Zefang Zong, Jingwei Wang, Yunke Zhang, Jingyi Wang, Xiaochong Lan, Jiahui Gong, Tianjian Ouyang, Fanjin Meng, et~al.
\newblock Towards large reasoning models: A survey of reinforced reasoning with large language models.
\newblock {\em arXiv preprint arXiv:2501.09686}, 2025.

\bibitem{DBLP:journals/corr/abs-2409-18807}
Zhuocheng Shen.
\newblock {LLM} with tools: {A} survey.
\newblock {\em CoRR}, abs/2409.18807, 2024.

\bibitem{DBLP:journals/csur/QinHLCDCZZHXHFSWQTZLSXZ25}
Yujia Qin, Shengding Hu, Yankai Lin, Weize Chen, Ning Ding, Ganqu Cui, Zheni Zeng, Xuanhe Zhou, Yufei Huang, Chaojun Xiao, Chi Han, Yi~R. Fung, Yusheng Su, Huadong Wang, Cheng Qian, Runchu Tian, Kunlun Zhu, Shihao Liang, Xingyu Shen, Bokai Xu, Zhen Zhang, Yining Ye, Bowen Li, Ziwei Tang, Jing Yi, Yuzhang Zhu, Zhenning Dai, Lan Yan, Xin Cong, Yaxi Lu, Weilin Zhao, Yuxiang Huang, Junxi Yan, Xu~Han, Xian Sun, Dahai Li, Jason Phang, Cheng Yang, Tongshuang Wu, Heng Ji, Guoliang Li, Zhiyuan Liu, and Maosong Sun.
\newblock Tool learning with foundation models.
\newblock {\em {ACM} Comput. Surv.}, 57(4):101:1--101:40, 2025.

\bibitem{DBLP:conf/nips/YangSLZGLS23}
Rui Yang, Lin Song, Yanwei Li, Sijie Zhao, Yixiao Ge, Xiu Li, and Ying Shan.
\newblock Gpt4tools: Teaching large language model to use tools via self-instruction.
\newblock In Alice Oh, Tristan Naumann, Amir Globerson, Kate Saenko, Moritz Hardt, and Sergey Levine, editors, {\em Advances in Neural Information Processing Systems 36: Annual Conference on Neural Information Processing Systems 2023, NeurIPS 2023, New Orleans, LA, USA, December 10 - 16, 2023}, 2023.

\bibitem{DBLP:conf/naacl/QiaoGLJC024}
Shuofei Qiao, Honghao Gui, Chengfei Lv, Qianghuai Jia, Huajun Chen, and Ningyu Zhang.
\newblock Making language models better tool learners with execution feedback.
\newblock In Kevin Duh, Helena G{\'{o}}mez{-}Adorno, and Steven Bethard, editors, {\em Proceedings of the 2024 Conference of the North American Chapter of the Association for Computational Linguistics: Human Language Technologies (Volume 1: Long Papers), {NAACL} 2024, Mexico City, Mexico, June 16-21, 2024}, pages 3550--3568. Association for Computational Linguistics, 2024.

\bibitem{deepresearch}
OpenAI.
\newblock Introducing deep research, 2025.

\bibitem{DBLP:conf/acl/TrivediBKS23}
Harsh Trivedi, Niranjan Balasubramanian, Tushar Khot, and Ashish Sabharwal.
\newblock Interleaving retrieval with chain-of-thought reasoning for knowledge-intensive multi-step questions.
\newblock In Anna Rogers, Jordan~L. Boyd{-}Graber, and Naoaki Okazaki, editors, {\em Proceedings of the 61st Annual Meeting of the Association for Computational Linguistics (Volume 1: Long Papers), {ACL} 2023, Toronto, Canada, July 9-14, 2023}, pages 10014--10037. Association for Computational Linguistics, 2023.

\bibitem{DBLP:conf/emnlp/ShaoGSHDC23}
Zhihong Shao, Yeyun Gong, Yelong Shen, Minlie Huang, Nan Duan, and Weizhu Chen.
\newblock Enhancing retrieval-augmented large language models with iterative retrieval-generation synergy.
\newblock In Houda Bouamor, Juan Pino, and Kalika Bali, editors, {\em Findings of the Association for Computational Linguistics: {EMNLP} 2023, Singapore, December 6-10, 2023}, pages 9248--9274. Association for Computational Linguistics, 2023.

\bibitem{DBLP:journals/corr/abs-2501-05366}
Xiaoxi Li, Guanting Dong, Jiajie Jin, Yuyao Zhang, Yujia Zhou, Yutao Zhu, Peitian Zhang, and Zhicheng Dou.
\newblock Search-o1: Agentic search-enhanced large reasoning models.
\newblock {\em CoRR}, abs/2501.05366, 2025.

\bibitem{DBLP:journals/corr/abs-2503-19470}
Mingyang Chen, Tianpeng Li, Haoze Sun, Yijie Zhou, Chenzheng Zhu, Haofen Wang, Jeff~Z. Pan, Wen Zhang, Huajun Chen, Fan Yang, Zenan Zhou, and Weipeng Chen.
\newblock Research: Learning to reason with search for llms via reinforcement learning.
\newblock {\em CoRR}, abs/2503.19470, 2025.

\bibitem{feng2025retoolreinforcementlearningstrategic}
Jiazhan Feng, Shijue Huang, Xingwei Qu, Ge~Zhang, Yujia Qin, Baoquan Zhong, Chengquan Jiang, Jinxin Chi, and Wanjun Zhong.
\newblock Retool: Reinforcement learning for strategic tool use in llms, 2025.

\bibitem{zheng2025deepresearcherscalingdeepresearch}
Yuxiang Zheng, Dayuan Fu, Xiangkun Hu, Xiaojie Cai, Lyumanshan Ye, Pengrui Lu, and Pengfei Liu.
\newblock Deepresearcher: Scaling deep research via reinforcement learning in real-world environments, 2025.

\bibitem{trivedi2021musique}
Harsh Trivedi, Niranjan Balasubramanian, Tushar Khot, and Ashish Sabharwal.
\newblock {M}u{S}i{Q}ue: Multihop questions via single-hop question composition.
\newblock {\em Transactions of the Association for Computational Linguistics}, 2022.

\bibitem{DBLP:conf/iclr/HendrycksBBZMSS21}
Dan Hendrycks, Collin Burns, Steven Basart, Andy Zou, Mantas Mazeika, Dawn Song, and Jacob Steinhardt.
\newblock Measuring massive multitask language understanding.
\newblock In {\em 9th International Conference on Learning Representations, {ICLR} 2021, Virtual Event, Austria, May 3-7, 2021}. OpenReview.net, 2021.

\bibitem{DBLP:journals/corr/abs-2402-03300}
Zhihong Shao, Peiyi Wang, Qihao Zhu, Runxin Xu, Junxiao Song, Mingchuan Zhang, Y.~K. Li, Y.~Wu, and Daya Guo.
\newblock Deepseekmath: Pushing the limits of mathematical reasoning in open language models.
\newblock {\em CoRR}, abs/2402.03300, 2024.

\bibitem{DBLP:conf/nips/LewisPPPKGKLYR020}
Patrick Lewis, Ethan Perez, Aleksandra Piktus, Fabio Petroni, Vladimir Karpukhin, Naman Goyal, Heinrich K{\"{u}}ttler, Mike Lewis, Wen{-}tau Yih, Tim Rockt{\"{a}}schel, Sebastian Riedel, and Douwe Kiela.
\newblock Retrieval-augmented generation for knowledge-intensive {NLP} tasks.
\newblock In Hugo Larochelle, Marc'Aurelio Ranzato, Raia Hadsell, Maria{-}Florina Balcan, and Hsuan{-}Tien Lin, editors, {\em Advances in Neural Information Processing Systems 33: Annual Conference on Neural Information Processing Systems 2020, NeurIPS 2020, December 6-12, 2020, virtual}, 2020.

\bibitem{DBLP:conf/iclr/AsaiWWSH24}
Akari Asai, Zeqiu Wu, Yizhong Wang, Avirup Sil, and Hannaneh Hajishirzi.
\newblock Self-rag: Learning to retrieve, generate, and critique through self-reflection.
\newblock In {\em The Twelfth International Conference on Learning Representations, {ICLR} 2024, Vienna, Austria, May 7-11, 2024}. OpenReview.net, 2024.

\bibitem{DBLP:conf/iclr/YoranWRB24}
Ori Yoran, Tomer Wolfson, Ori Ram, and Jonathan Berant.
\newblock Making retrieval-augmented language models robust to irrelevant context.
\newblock In {\em The Twelfth International Conference on Learning Representations, {ICLR} 2024, Vienna, Austria, May 7-11, 2024}. OpenReview.net, 2024.

\bibitem{DBLP:conf/emnlp/Yang0ZBCSM18}
Zhilin Yang, Peng Qi, Saizheng Zhang, Yoshua Bengio, William~W. Cohen, Ruslan Salakhutdinov, and Christopher~D. Manning.
\newblock Hotpotqa: {A} dataset for diverse, explainable multi-hop question answering.
\newblock In Ellen Riloff, David Chiang, Julia Hockenmaier, and Jun'ichi Tsujii, editors, {\em Proceedings of the 2018 Conference on Empirical Methods in Natural Language Processing, Brussels, Belgium, October 31 - November 4, 2018}, pages 2369--2380. Association for Computational Linguistics, 2018.

\bibitem{DBLP:journals/corr/abs-2110-14168}
Karl Cobbe, Vineet Kosaraju, Mohammad Bavarian, Mark Chen, Heewoo Jun, Lukasz Kaiser, Matthias Plappert, Jerry Tworek, Jacob Hilton, Reiichiro Nakano, Christopher Hesse, and John Schulman.
\newblock Training verifiers to solve math word problems.
\newblock {\em CoRR}, abs/2110.14168, 2021.

\bibitem{deepseekai2025deepseekr1incentivizingreasoningcapability}
DeepSeek-AI.
\newblock Deepseek-r1: Incentivizing reasoning capability in llms via reinforcement learning, 2025.

\bibitem{DBLP:conf/eurosys/ShengZYWZZPL025}
Guangming Sheng, Chi Zhang, Zilingfeng Ye, Xibin Wu, Wang Zhang, Ru~Zhang, Yanghua Peng, Haibin Lin, and Chuan Wu.
\newblock Hybridflow: {A} flexible and efficient {RLHF} framework.
\newblock In {\em Proceedings of the Twentieth European Conference on Computer Systems, EuroSys 2025, Rotterdam, The Netherlands, 30 March 2025 - 3 April 2025}, pages 1279--1297. {ACM}, 2025.

\bibitem{DBLP:journals/corr/abs-2405-13576}
Jiajie Jin, Yutao Zhu, Xinyu Yang, Chenghao Zhang, and Zhicheng Dou.
\newblock Flashrag: {A} modular toolkit for efficient retrieval-augmented generation research.
\newblock {\em CoRR}, abs/2405.13576, 2024.

\bibitem{yang2024qwen2}
An~Yang, Baosong Yang, Beichen Zhang, Binyuan Hui, Bo~Zheng, Bowen Yu, Chengyuan Li, Dayiheng Liu, Fei Huang, Haoran Wei, et~al.
\newblock Qwen2. 5 technical report.
\newblock {\em arXiv preprint arXiv:2412.15115}, 2024.

\bibitem{DBLP:journals/nature/MnihKSRVBGRFOPB15}
Volodymyr Mnih, Koray Kavukcuoglu, David Silver, Andrei~A. Rusu, Joel Veness, Marc~G. Bellemare, Alex Graves, Martin~A. Riedmiller, Andreas Fidjeland, Georg Ostrovski, Stig Petersen, Charles Beattie, Amir Sadik, Ioannis Antonoglou, Helen King, Dharshan Kumaran, Daan Wierstra, Shane Legg, and Demis Hassabis.
\newblock Human-level control through deep reinforcement learning.
\newblock {\em Nat.}, 518(7540):529--533, 2015.

\bibitem{DBLP:conf/icml/WangSHHLF16}
Ziyu Wang, Tom Schaul, Matteo Hessel, Hado van Hasselt, Marc Lanctot, and Nando de~Freitas.
\newblock Dueling network architectures for deep reinforcement learning.
\newblock In Maria{-}Florina Balcan and Kilian~Q. Weinberger, editors, {\em Proceedings of the 33nd International Conference on Machine Learning, {ICML} 2016, New York City, NY, USA, June 19-24, 2016}, volume~48 of {\em {JMLR} Workshop and Conference Proceedings}, pages 1995--2003. JMLR.org, 2016.

\bibitem{DBLP:journals/corr/ThomasB17}
Philip~S. Thomas and Emma Brunskill.
\newblock Policy gradient methods for reinforcement learning with function approximation and action-dependent baselines.
\newblock {\em CoRR}, abs/1706.06643, 2017.

\bibitem{DBLP:conf/icml/MnihBMGLHSK16}
Volodymyr Mnih, Adri{\`{a}}~Puigdom{\`{e}}nech Badia, Mehdi Mirza, Alex Graves, Timothy~P. Lillicrap, Tim Harley, David Silver, and Koray Kavukcuoglu.
\newblock Asynchronous methods for deep reinforcement learning.
\newblock In Maria{-}Florina Balcan and Kilian~Q. Weinberger, editors, {\em Proceedings of the 33nd International Conference on Machine Learning, {ICML} 2016, New York City, NY, USA, June 19-24, 2016}, volume~48 of {\em {JMLR} Workshop and Conference Proceedings}, pages 1928--1937. JMLR.org, 2016.

\bibitem{DBLP:journals/corr/SchulmanWDRK17}
John Schulman, Filip Wolski, Prafulla Dhariwal, Alec Radford, and Oleg Klimov.
\newblock Proximal policy optimization algorithms.
\newblock {\em CoRR}, abs/1707.06347, 2017.

\bibitem{DBLP:conf/nips/RafailovSMMEF23}
Rafael Rafailov, Archit Sharma, Eric Mitchell, Christopher~D. Manning, Stefano Ermon, and Chelsea Finn.
\newblock Direct preference optimization: Your language model is secretly a reward model.
\newblock In Alice Oh, Tristan Naumann, Amir Globerson, Kate Saenko, Moritz Hardt, and Sergey Levine, editors, {\em Advances in Neural Information Processing Systems 36: Annual Conference on Neural Information Processing Systems 2023, NeurIPS 2023, New Orleans, LA, USA, December 10 - 16, 2023}, 2023.

\bibitem{zhang2025grpoleaddifficultyawarereinforcementlearning}
Jixiao Zhang and Chunsheng Zuo.
\newblock Grpo-lead: A difficulty-aware reinforcement learning approach for concise mathematical reasoning in language models, 2025.

\bibitem{DBLP:journals/corr/abs-2503-21460}
Junyu Luo, Weizhi Zhang, Ye~Yuan, Yusheng Zhao, Junwei Yang, Yiyang Gu, Bohan Wu, Binqi Chen, Ziyue Qiao, Qingqing Long, Rongcheng Tu, Xiao Luo, Wei Ju, Zhiping Xiao, Yifan Wang, Meng Xiao, Chenwu Liu, Jingyang Yuan, Shichang Zhang, Yiqiao Jin, Fan Zhang, Xian Wu, Hanqing Zhao, Dacheng Tao, Philip~S. Yu, and Ming Zhang.
\newblock Large language model agent: {A} survey on methodology, applications and challenges.
\newblock {\em CoRR}, abs/2503.21460, 2025.

\bibitem{DBLP:journals/corr/abs-2309-14365}
Pengyu Zhao, Zijian Jin, and Ning Cheng.
\newblock An in-depth survey of large language model-based artificial intelligence agents.
\newblock {\em CoRR}, abs/2309.14365, 2023.

\bibitem{DBLP:journals/tmlr/SumersYN024}
Theodore~R. Sumers, Shunyu Yao, Karthik Narasimhan, and Thomas~L. Griffiths.
\newblock Cognitive architectures for language agents.
\newblock {\em Trans. Mach. Learn. Res.}, 2024, 2024.

\bibitem{AutoAgent}
Chao~Huang Jiabin~Tang, Tianyu~Fan.
\newblock {AutoAgent: A Fully-Automated and Zero-Code Framework for LLM Agents}, 2025.

\bibitem{DBLP:journals/corr/abs-2306-02224}
Hui Yang, Sifu Yue, and Yunzhong He.
\newblock Auto-gpt for online decision making: Benchmarks and additional opinions.
\newblock {\em CoRR}, abs/2306.02224, 2023.

\bibitem{langchain}
Langchain, 2023.

\end{thebibliography}

\end{document}